\documentclass{article}

\usepackage[preprint]{neurips_wrl2020}

\usepackage[utf8]{inputenc} % allow utf-8 input
\usepackage[T1]{fontenc}    % use 8-bit T1 fonts
\usepackage{hyperref}       % hyperlinks
\usepackage{url}            % simple URL typesetting
\usepackage{booktabs}       % professional-quality tables
\usepackage{amsfonts}       % blackboard math symbols
\usepackage{nicefrac}       % compact symbols for 1/2, etc.
\usepackage{microtype}      % microtypography

\usepackage{amssymb,graphicx}
\usepackage{subcaption}
\usepackage{wrapfig}

\usepackage{amsmath}

\title{Semantics and explanation: why counterfactual explanations produce adversarial examples in deep neural networks}

\author{
  Kieran Browne\thanks{Corresponding Author} \\
  Research School  of Humanities \& the Arts\\
  Australian National University\\
  \texttt{kieran.browne@anu.edu.au} \\
   \And
  Ben Swift \\
  Research School of Computer Science\\
  Australian National University\\
  \texttt{ben.swift@anu.edu.au} \\
}

\begin{document}

\maketitle

\begin{abstract}
  Recent papers in explainable AI have made a compelling case for counterfactual modes of explanation. While counterfactual explanations appear to be extremely effective in some instances, they are formally equivalent to adversarial examples. This presents an apparent paradox for explainability researchers: if these two procedures are formally equivalent, what accounts for the explanatory divide apparent between counterfactual explanations and adversarial examples? We resolve this paradox by placing emphasis back on the semantics of counterfactual expressions. Producing satisfactory explanations for deep learning systems will require that we find ways to interpret the semantics of hidden layer representations in deep neural networks.
\end{abstract}

%\keywords{explainable AI \and counterfactual explanation \and adversarial examples \and semantics}

\section{Introduction}

Deep neural networks (DNNs) will not be explainable without first addressing the scarcity of semantics. Computational methods already exist to produce model-agnostic explanations that are understandable to laypersons. These methods simply do not function as explanations when applied to ambiguous or low-level representations that are common to DNNs. We will argue that this is not simply a limitation of existing explanatory methods, but rather that \emph{there can be no explanation without semantics}. Because deep learning (DL) typically operates on ``raw data'', with little semantic content (e.g. pixels and characters), this realisation serves to clarify the explainability challenge; we either find a way to extract the semantics presumed to exist in the hidden layers of the network or concede defeat.

Recent papers in explainable artificial intelligence (XAI) have identified problems with the field's theoretical bases. \citet{miller2019} argues that the field typically operates with only an intuitive notion of what explanation is; and one which is divorced from how humans explain and understand explanation. He proposes that XAI adopt ``everyday explanations'', based on a set of principles from psychological and social scientific research. Similarly, \citet{wachter2018} propose ``counterfactual explanations'' which are consistent with the principles identified by Miller. Wachter et al. additionally specify a method for generating counterfactual explanations. Counterfactual explanations, as Wachter et al. demonstrate, are model-agnostic, automatically computable and comprehensible to laypersons. The authors argue that these counterfactual explanations offer the path to explaining complex algorithmic systems to anyone. However, equivalent computations have been used in DL research since 2014, though not to produce explanations. Instead, in the context of DL research, the counterfactual computation produces ``adversarial examples''; imperceptibly modified inputs which cause the network to inexplicably and confidently misclassify. 

This should give us pause for thought; how is it possible that the same method can on the one hand represent a promising new means of explaining the decisions of a DNN to anyone, and on the other hand represent a confounding brittleness in that same decision making process? We call this phenomenon, \emph{the explanatory divide}. We will argue that this divide reveals a blind spot in XAI research with regards to semantics.

\subsection{The argument}

This article proceeds as follows: we begin in Section \ref{background} by outlining the history of XAI research and its revival in the era of deep learning. In Section \ref{why-now} we describe the novel social challenge posed by this technology and how this has affected the problem of explainability. In Section \ref{human-turn} we introduce two recent papers that have challenged the prevailing methods in XAI by introducing human-centric modes of explanation to the field. In Section \ref{maths} we show the equivalence of counterfactual methods proposed by Wachter et al. with those used to generate adversarial examples and examine the explanatory divide apparent in the two usages. In Section \ref{wachters-account} we refute Wachter et al.'s account of the explanatory divide. In Section \ref{semantics} we argue that the explanatory divide is instead a consequence of the semantic content of the perturbed vector. In Section \ref{semantics-dl} we show that semantics is a blind spot in XAI research attributable to researchers' concern for a computational solution. In Section \ref{dl-culture} we argue that semantic issues are endemic to DL due to operating on ``raw data''. In Section \ref{semantic-challenges} we examine the existing research in extracting the semantics of hidden layers and discuss the ongoing challenges. We conclude by proposing a possible path forward for combining existing explainability methods with the partially known semantics of DNNs.

\subsection{A note on ``semantics''}\label{terms}

Although ``semantics'' is a common term in DL literature, its use is ambiguous. In the early literature on DL it is commonly claimed that DNNs learn semantic features automatically in order to solve problems (e.g. \cite{bengio2009}, \citet[p. 441]{lecun2015}). This is based on the assumption that in order to solve complex problems like image classification, the network must generate intelligible intermediate representations (e.g. whiskers and paws used to identify cats). However this kind of semantics has not been reliably shown to exist and it remains a significant challenge to find a mapping between the latent spaces of DNNs and human concepts. Other researchers appear to use ``semantics'' to refer to any kind of internal representation, that is, any way of carving up the world whether or not it maps to something a human might understand. These duplicate uses of ``semantics'' cause significant ambiguity and some authors have resorted to tautology to distinguish the two. \cite{biran2017} for example use ``semantically meaningful representations'' to distinguish internal representations which correspond to categories which humans (or perhaps specifically English speakers) find meaningful.

Following the language of semiotics (see \cite{morris1938}), we take semantics to be the relation between sign and signified. This is distinguished from syntactics (relations between signs) and pragmatics (the relation between signs and the interpreter). Of course the representations in DNNs are not really signs, at least not in the standard sense. Their relation to meanings are correlative and continuous rather than discrete as in symbolic systems. Although some accounts of meaning disallow fuzzy concepts, others (e.g. later Wittgetnstein) argue that many of our concepts have ``blurred edges'' and we are able to use them productively nonetheless~\cite[sect. 71]{PI}.
 This is the sense in which we suggest ``semantics'' should be understood in DL. Whether we consider a hidden unit to \emph{mean} ``whiskers'' then, depends on how reliably it correlates to the English language concept ``whiskers''.

\section{Background}\label{background}

DL has afforded significant advances in a broad range of problems. However, little progress has been made in explaining the behaviour and decision-making processes of these systems. Although the reinterpretation of machine learning as artificial intelligence in the 1990s revived the decades-old field of XAI (eXplainable Artificial Intelligence), DL remains a ``black box''---a descriptor that has followed DNNs and their precursory methods since the 1990s \citep{spining1994}\citep{adadi2018}.

This desire for explanations of algorithmic decisions predates DL, beginning in the context of rule-based expert systems as early as the 1970s \citep{biran2017}. Research on XAI has been tied to AI such that it has endured the same periods of disenchantment known colloquially as ``AI winters''. An era-agnostic survey of explainability is provided in \cite{biran2017}. We will focus our account on the contemporary (deep) neural network paradigm of AI and XAI. 

Since the rise of DL in the mid 2000s, artificial neural networks (ANNs) have become more complex by orders of magnitude. Unlike simpler statistical models, ANNs are generally considered to be ``black boxes'' because the representations they generate are not readily interpretable \citep{breiman2001}.

In machine learning, the goal of explainability has often been pursued through visualisation \citep{biran2017}. In late 80s and early 90s, a number of diagrammatic visualisations emerged for ANNs; these usually relied on a traditional graph-theoretic ``nodes and edges'' representation augmented with edge-weight information. As networks increased in size throughout the 90s these images became increasingly difficult to interpret \citep{browne2018critical}. For deep architectures of contemporary scale they are essentially obsolete e.g. Microsoft's Turing Natural Language Generation T-NLG has 17 billion parameters~\citep{rossetTuringNLG17billionparameterLanguage2020}. Today, DL visualisations tend to represent only single layers or single neurons rather than an entire network~\citep{olah2017}.

The other common approach in explainability research is to approximate the behaviour of an ANN with a more ``interpretable'' model. In the late 80s early 90s this was usually referred to as \emph{rule extraction} \citep{andrews1995}. This meant distilling the many calculations of a neural network into a series \texttt{IF...THEN} rules akin to symbolic AI. Rule extraction is rarely mentioned in the ``deep'' era of neural networks, but similar methods are still used under alternative names such as \emph{knowledge distillation} \citep{hinton2015}, which notably omits any reference to explainability. Knowledge distillation, like rule extraction, uses a trained DNN to train a simpler, \emph{interpretable} model such as a decision tree \citep{frosst2017}. From an XAI perspective a lingering conceptual problem remains; if the simpler model is similar enough to capture the decisions of the DNN, why use DL at all?

\subsection{Why now?}\label{why-now}

It is still common to see XAI papers use adoption as a motivation for explainability. If users do not understand or trust the model, we are told, they will choose not to use it \citep{ribeiro2016}. While this justification has been made many times, it has little to do with the need for explainability as it exists today. While some opt-in AI-branded DL services do exist, the significant growth area for DL is in systems that people are \emph{subject to} through institutional means. The application of DL in institutions of social and political importance; e.g. banks, courts, media distribution, political campaigning etc. has naturally drawn increasing attention from social scientists and increased scrutiny from law makers \citep{barocas2016}\citep{burrell2016}\citep{stilgoe2018}.

Explainability matters now more than ever because DL is being used to determine social realities, e.g. by banks to distribute credit or by the justice system to decide parole. Here we are in danger of conflating \emph{prediction} with \emph{prescription}. To borrow the language of speech acts, prediction is \emph{constative}; that is, it makes a claim about the world, e.g. ``the house price will be \$1,000,000'', ``this is a picture of a tennis ball'' or ``this digit is a 6'', which may be evaluated by independent observation as more or less accurate.
In normal use, the truth is independent of the prediction, and the predictions may be judged as more or less accurate. However, used prescriptively, e.g. job applications, loan decisions, bail decisions, social reality is wholly determined by the prediction. In these cases the need for an explanation of the network's outputs/outcomes is paramount---there is no independent ground truth outside the algorithmic decision.

\subsection{A human turn in explainable AI?}\label{human-turn} 

XAI has been significantly siloed from other disciplinary understandings of explanation. However two recent papers propose new approaches sensitive to the human factors of explanation. Both draw on bodies of knowledge from outside computer science to propose modes of explanation inspired by human-to-human explanations. The first we will discuss, from XAI researcher \citet{miller2019}, draws on research from philosophy, psychology, social science and cognitive science to provide a theoretical framework for XAI sensitive to how humans explain and understand explanation. The second, from an interdisciplinary team led by \citet{wachter2018}, proposes a practical method for providing a legally-compelled explanation for those subject to algorithmic decision making. These two approaches appear to have emerged independently but are complementary. Both argue for a shift in explainability research toward social and context-dependent modes of explanation.

Miller's \emph{everyday explanations} provide a conceptual foundation for this development. Raising a concern about the theoretical underpinnings of XAI, he claims that most research is guided merely by researchers' ``intuition'' for what constitutes a good explanation and argues that computational solutions are not sufficient for explainability \citep{miller2019}. Miller argues that XAI should take inspiration from the way humans explain to each other. He surveys existing literature on explanation in philosophy, psychology, social science and cognitive science in order to draw four conclusions about explanations:

1. Explanations are \emph{contrastive}; that is, they ``explain the cause of an event \emph{relative to some other event} that did not occur''.

2. They are \emph{selective}; that is, we rarely if ever give an explanation that describes the ``complete'' cause of an event.

3. They are \emph{social}; that is, they are presented relative to who the explainee is, and what they can be expected to understand.

4. \emph{Probabilities probably don't matter}; that is, statistical explanations of events are unsatisfying unless accompanied by causal explanations.

Around the same time, Wachter et al.'s \emph{counterfactual explanations} appeared in the \emph{Harvard Journal of Law \& Technology} \citep{wachter2018}, motivated by the looming challenge of the ``right to an explanation'' under the European Union's General Data Protection Regulation's (GDPR) and the competing technological, social and legal challenges therein. The paper proposes the counterfactual explanation as a way to offer meaningful explanations of algorithmic decisions to those affected. Counterfactual explanations are a model-agnostic method for generating explanations of algorithmic decisions for a lay audience based on the notion of the \emph{counterfactual} from the philosophy of causation.

A counterfactual explanation is defined by Wachter et al. as a minimal set of changes to the input data found to produce a desired decision in the network.

More formally:

\begin{quote}
``Score $p$ was returned because variables $V$ had values $(v_1, v_2,...)$
associated with them. If $V$ instead had values $(v_1', v_2',...)$, and all
other variables had remained constant, score $p'$ would have been returned.''
\end{quote}

Wachter et al. go on to present a small number of case studies which demonstrate the efficacy of this method in real-world cases. The method is extremely effective in the scenarios described. When applied to a specific example, as in the paper's example of a bank loan, the above formal definition is translated into something which reads as plain English:

\begin{quote}
``You were denied a loan because your annual income was \pounds 30,000. If your
income had been \pounds45,000, you would have been offered a loan.''
\end{quote}

The counterfactual explanation embodies the principles of everyday explanations articulated in Miller's paper. Where existing research treats explanation axiomatically, counterfactuals are
conscious of the audience, i.e. they are \emph{social}. The counterfactual explanation is for a lay audience, specifically the GDPR's ``data subjects''. Counterfactual explanations are also \emph{selective}; as the title of the paper suggests, they allow for explanations ``without opening the black box,'' or in other words, without completely revealing how the algorithm works. Most strikingly, the counterfactual explanation is \emph{contrastive}; it points to the changes in the input which would have resulted in an alternative outcome.

\section{Adversarial examples are counterfactual explanations}\label{maths}

The counterfactual explanation is only nominally new to DL/XAI. Since 2014 the generative perturbation of input vectors to probe at decision boundaries has been the topic of a significant body of research under the banner of ``adversarial examples'' \citep{szegedy2014}\citep{nguyen2015}\citep{moosavi-dezfooli2016}\citep{su2019}. Adversarial examples, like counterfactual explanations, are algorithmically generated perturbations to input data which are optimised to alter the DNN's output in a particular manner. Much of this research has concerned itself with image classification, such as convolutional neural networks, however, successful adversarial examples have been produced in audio \citep{alzantot2018} and text \citep{papernot2016} domains as well.

In adversarial examples, imperceptible (to a human observer) changes to the input cause the network to entirely and confidently misclassify. The originators of this line of research, \citet{szegedy2014} call this phenomenon ``intriguing'' and ``counter-intuitive'' while later researchers have commonly described the adversarial example in terms of a vulnerability to attack \citep{kurakin2016}\citep{su2019}.

A more formal definition helps to clarify the equivalence between adversarial examples and counterfactual explanations. Both are defined as a constrained optimisation problem where the objective is to change the network's output to a some other output by minimally altering the input. Consider a DL classifier

\begin{equation}
f_w(x) = y
\end{equation}

where $y$ is the predicted class of input $x$. For both adversarial examples and counterfactual explanations we seek an input $x'$ as close as possible to $x$ such that our network $f_w$ classifies $x'$ as a different target class $y'$. This can be written as an optimisation problem:

\begin{equation}
\begin{aligned}
& \underset{x'}{arg\,min} & & d(x, x') \\
& \text{subject to}
& & f_w(x') = y' \neq y \\
\end{aligned}
\end{equation}

The distance metric $d$ is measure of the distortion of $x'$ relative to $x$---the distance between the original input and the altered one. The objective is produce the target output $y'$ while minimising the distance $d(x, x')$. As a consequence the definition of $d$ will influence the resulting of adversarial or counterfactual.

The first paper to propose adversarial examples uses the Euclidian distance ($L_2$ norm):

\begin{equation}
d(x, x') =  \lvert x' - x \rvert_2 
\end{equation}

Later papers including Wachter et al.'s use other metrics, including the Manhattan distance ($L_1$ norm), but these metrics do not fundamentally change the nature of the method.

This presents an apparent paradox for explainability researchers: if these two procedures are formally equivalent, what accounts for the explanatory divide apparent between counterfactual explanations and adversarial examples?

\subsection{Making sense of the explanatory divide}\label{wachters-account} 

\citet{wachter2018} acknowledge that an adversarial example is ``a counterfactual by a different name,'' but appear unconcerned by this, and propose two grounds for the explanatory divide (our term). The first is that ``none of the standard works on adversarial perturbations make use of appropriate distance functions'' and the second is that adversarial examples are invalid because they do not come from the ``space of real-images'' and therefore do not qualify as ``possible worlds''. The first amounts to a challenge over the ``correct'' definition of the distance metric $d$ in Equation 3, the second is a metaphysical claim.

The claim that none of the ``standard works'' on adversarial examples use an appropriate distance metric needs to be understood in the context of Wachter et al.'s own discussion of the properties of an appropriate distance metric. While they stress that case-specific considerations must be taken into account, they suggest as a first approximation to use the $L_1$ norm weighted by the inverse median absolute deviation (MAD). The MAD is chosen for its robustness to outliers, while the $L_1$ norm is chosen for its sparsity-inducing properties; i.e. it restricts differences to as few input dimensions as possible.

From these properties it is possible to understand the concern the authors have with the distance metrics favoured by the adversarial example research community. Wachter et al. are correct that the majority of adversarial example research ``favour[s] making small changes to many variables'' so that the difference is diluted across the inputs and this contributes to the indistinguishableness of the perturbations. We agree that this would theoretically make these less useful as explanations; in particular that it contravenes Miller's principle that explanations should be \emph{selective}. However, although sparse counterfactuals may be preferable to dense ones, sparsity is not in itself sufficient for explanation. This is clear from the research of \citet{su2019}, who do restrict their variations to a single input feature, in this case a single pixel, whose impact is visually identifiable.

Su et al. demonstrate that in the majority of cases, changing a single pixel is enough to cause a network to misclassify to at least one other class. The generated perturbations are sparse and salient and in spite of this, they remain distinctly adversarial.

Wachter et al.'s second account for the explanatory divide is that adversarially-perturbed images do not represent ``possible worlds''. The intuition here is that standard adversarial perturbations appear as very slight ``noise'' are distinctly \emph{not} random; instead they encode the signature of a class that is not present in ``natural'' images. Perhaps there is some truth to this---it does seem unlikely that noise with these very specific properties would occur by chance. However, there is something distinctly unsatisfying about this account. Should we regard crafted or manipulated images as ``impossible''? We live amongst a proliferation of unnatural images. Additionally, adversarially-perturbed images have been shown to work in the real world even when printed out and photographed through low-quality cameras \citep{kurakin2016}\citep{brown2017}\citep{jan2019}. Whether or not adversarial examples are ``possible'' without contrivance, researchers must take seriously the possibility of encountering adversarial examples that have been intentionally planted in the world.

\subsection{The explanatory divide and semantics}\label{semantics} 

If the explanatory divide cannot be accounted for by poor distance metrics or impossible worlds, how else can we make sense of it? We believe the answer lies in the semantics of the perturbed vector.

An example helps to clarify this assertion. The input data to the AI decision processes in Wachter et al.'s examples are expressed using semantically dense and contextually relevant dimensions: income, grade-point average, body-mass index, etc. Some of these also represent factors that we (humans) might consider appropriate evidence to base a loan decision on, others are certainly not (e.g. age, and race). Regardless, a counterfactual explanation that operates on semantically dense dimensions helps us to understand the decision even if it causes us to question its validity. A counterfactual for a hiring decision that identifies race, gender or age as deciding factors is explanatory even if it only provides a justification for disregarding the results. In the framework of Miller's everyday explanations, these dimensions are \emph{social} as they can be expected to be understood by the explainee.

In contrast, adversarial examples are produced when the same computation is applied to data with little semantic content. Much of DL operates on ``raw data'', i.e. individual pixels, letters, waveform samples, bits etc. Reductio ad absurdum; explaining an image classified as ``building'' based on the redness value in a particular pixel is unsurprisingly unhelpful. Instead, the factors that a human would consider to be relevant are dispersed and discontinuous in pixel space, \emph{they are not discoverable using sparse or dense perturbations}. Debiasing the network is also extremely challenging when operating with low-level semantics, because factors we would wish to disallow are equally dispersed and discontinuous. 

Mathematically speaking, there is no difference between a vector of pixel values and a vector of semantically rich features. Therefore, the crucial relationship from an XAI perspective is not between the network and the computation producing the explanation, but between the semantics in the network and the human explainee.

\section{Semantics is the core challenge of explaining deep learning}\label{semantics-dl} 

As we have discussed, the efforts of XAI researchers have been significantly focused on finding ways to reduce the complexity of a given network. These methods share the same fatal flaw as the counterfactual, no computation can get around the semantic problem. Any explanatory technique will produce a non-sequitur if it, for example, attempts to explain driving instructions from pixel values. For XAI this appears to be a significant blindspot.

\subsection{The culture of Deep Learning}\label{dl-culture}

Of course, the simplest solution to this explainability problem would be to apply DL only to higher-level, contextually relevant representations. But this would require us to forego what \citet{lecun2015} call the ``key advantage'' of DNNs; that they can operate on ``raw data'' and do not require feature engineering to produce useful results.

Whether counterfactual explanations or any other one of a trove of abandoned methods (rule-extraction, random-forests, etc.) may again be used to explain DNNs rests on whether the network's learned semantics, in the sense we define in Section \ref{terms}, can be discovered.

\subsection{In search of semantics}\label{semantic-challenges} 

Although significant early papers in the DL literature presume that DNNs discover their own semantics in order to solve problems \citep{bengio2009}\citep{lecun2015} the community as a whole appears to have quietly abandoned this assertion in recent years. We believe this to be a fatal error if we hope to explain these systems.

Let us assume for a moment that semantics can be discovered.
Given a clear knowledge of the semantics of hidden layer neurons in a network, it would be possible to generate counterfactual explanations consisting only of semantically dense and contextually relevant dimensions in the network's feature space, perhaps even without needing to synthesise inputs in pixel space. A counterfactual explanation at this level might read:

\begin{quote}
The input image was labelled ``building'' because hidden neuron 41435, which generally activates for hubcaps, had an activation of $0.32$. If hidden neuron 41435 had an activation of $0.87$ the input image would have been labelled ``car''.
\end{quote}

This is a contrived example, but it illustrates what should be possible if the hidden represenations in DNNs were interpretable.
The problem is that we are either yet to develop appropriate tools to discover these DL's internal semantics, or they do not exist.

\subsection{Revealing hidden layers} \label{revealing}

Researchers have made some progress in identifying the semantics of the hidden units (``neurons'') (see e.g. \cite{erhan2009}, \cite{olah2017}). In one of the earliest cases, \citet{erhan2009} identifies a neuron that appears to represent ``faces'', although it remains unclear how general/specific this category actually is (see \cite{browne2018critical}). A number of visualisation methods have been developed which serve to interpret hidden representations in DNNs. One technique is simply to collate dataset examples that maximise the activation of a given hidden unit \citep{szegedy2014}. Humans are often able to perceive commonalities between these high ranking examples. However, this method is susceptible to confounding factors. Saliency maps (see e.g. \citet{simonyan2014}, \citet{fong2017}) serve to avoid some of these pitfalls. They visualise the area(s) of images which contribute significantly to a hidden unit's activation, allowing a viewer to identify contributing visual features. Feature visualisations (e.g. \cite{erhan2009}, \cite{le2013}, \cite{nguyen2015}, \cite{karpathy2016}) synthesise images which maximally activate a hidden unit. \citet{olah2018} demonstrate that using many of these methods in tandem can be particularly enlightening.

Using methods described above, \citet{olah2017} discover a neuron that responds to different kinds of sports balls (e.g. golf balls, tennis balls, footballs, baseballs) (Fig. 2). This is particularly compelling because the neuron appears to have captured something approximating the human category that might be called ``sports ball'' in spite of their differing appearances and contexts. However, this is a particularly favourable example, as units with clear semantics appear to be exception and not the rule. Olah et al. found a number of cases where representations were ``poly-semantic'' \citep{olah2018} e.g. a unit was discovered that activates for cats and foxes but also cars. This is difficult to make sense of given that there is no apparent visual, contextual or categorical similarity between the cats and cars. In other instances
units appeared to have no discernible semantics whatsoever \citep{olah2017}. In summary, while there have been a number of very promising cases, the extraction of semantics from hidden units is far from a solved problem. 

Olah et al. note that single neurons (i.e. standard basis vectors) may not necessarily be the vector directions with a clear one-to-one mapping with English semantics, and that other basis vectors seem to be just as meaningful \citep{olah2017}. While this increases the likelihood of some arbitrary direction mapping to a human concept, it also makes the search space essentially infinite (floating point precision notwithstanding).

\section{Conclusion}\label{the-path-forward}

The equivalence of adversarial examples and counterfactual explanations demonstrates the necessity of semantics to the problem of explainability. Semantics appears to be a blindspot for XAI, which has instead focussed on computational innovations. The necessary computational methods to explain DL already exist---provided we use semantically rich and contextually relevant representations as inputs, or we can discover the semantics in hidden layers. With current research these semantics have not been shown consistently to exist and be discoverable.

\bibliography{bibliography}  % .bib

\begin{thebibliography}{35}
\providecommand{\natexlab}[1]{#1}
\providecommand{\url}[1]{\texttt{#1}}
\expandafter\ifx\csname urlstyle\endcsname\relax
  \providecommand{\doi}[1]{doi: #1}\else
  \providecommand{\doi}{doi: \begingroup \urlstyle{rm}\Url}\fi

\bibitem[Miller(2019)]{miller2019}
Tim Miller.
\newblock Explanation in artificial intelligence: Insights from the social
  sciences.
\newblock \emph{Artificial Intelligence}, 267:\penalty0 1--38, 2019.

\bibitem[Wachter et~al.(2018)Wachter, Mittelstadt, and Russell]{wachter2018}
Sandra Wachter, Brent Mittelstadt, and Chris Russell.
\newblock Counterfactual explanations without opening the black box: Automated
  decisions and the {GDPR}.
\newblock \emph{Harvard Journal of Law \& Technology}, 31\penalty0 (2), 2018.

\bibitem[Bengio et~al.(2009)]{bengio2009}
Yoshua Bengio et~al.
\newblock Learning deep architectures for ai.
\newblock \emph{Foundations and trends{\textregistered} in Machine Learning},
  2\penalty0 (1):\penalty0 1--127, 2009.

\bibitem[LeCun et~al.(2015)LeCun, Bengio, and Hinton]{lecun2015}
Yann LeCun, Yoshua Bengio, and Geoffrey Hinton.
\newblock Deep learning.
\newblock \emph{nature}, 521\penalty0 (7553):\penalty0 436--444, 2015.

\bibitem[Biran and Cotton(2017)]{biran2017}
Or~Biran and Courtenay Cotton.
\newblock Explanation and justification in machine learning: A survey.
\newblock In \emph{IJCAI-17 workshop on explainable AI (XAI)}, volume~8, 2017.

\bibitem[Morris(1938)]{morris1938}
Charles~William Morris.
\newblock Foundations of the theory of signs.
\newblock In \emph{International encyclopedia of unified science}, volume~1,
  pages 1--59. Chicago University Press, 1938.

\bibitem[Wittgenstein(1958)]{PI}
Ludwig Wittgenstein.
\newblock \emph{Philosophical investigations}.
\newblock Basil Blackwell Ltd, 1958.

\bibitem[Spining et~al.(1994)Spining, Darsey, Sumpter, and Nold]{spining1994}
MT~Spining, JA~Darsey, BG~Sumpter, and DW~Nold.
\newblock Opening up the black box of artificial neural networks.
\newblock \emph{Journal of chemical education}, 71\penalty0 (5):\penalty0 406,
  1994.

\bibitem[Adadi and Berrada(2018)]{adadi2018}
Amina Adadi and Mohammed Berrada.
\newblock Peeking inside the black-box: A survey on explainable artificial
  intelligence (xai).
\newblock \emph{IEEE Access}, 6:\penalty0 52138--52160, 2018.

\bibitem[Breiman(2001)]{breiman2001}
Leo Breiman.
\newblock Statistical modeling: The two cultures.
\newblock \emph{Statistical science}, 16\penalty0 (3):\penalty0 199--231, 2001.

\bibitem[Browne et~al.(2018)Browne, Swift, and Gardner]{browne2018critical}
Kieran Browne, Ben Swift, and Henry Gardner.
\newblock Critical challenges for the visual representation of deep neural
  networks.
\newblock In \emph{Human and Machine Learning}, pages 119--136. Springer, 2018.

\bibitem[Rosset(2020)]{rossetTuringNLG17billionparameterLanguage2020}
Corby Rosset.
\newblock Turing-{{NLG}}: {{A}} 17-billion-parameter language model by
  {{Microsoft}}, February 2020.
\newblock Library Catalog: www.microsoft.com.

\bibitem[Olah et~al.(2017)Olah, Mordvintsev, and Schubert]{olah2017}
Chris Olah, Alexander Mordvintsev, and Ludwig Schubert.
\newblock Feature visualization.
\newblock \emph{Distill}, 2017.
\newblock \doi{10.23915/distill.00007}.

\bibitem[Andrews et~al.(1995)Andrews, Diederich, and Tickle]{andrews1995}
Robert Andrews, Joachim Diederich, and Alan~B Tickle.
\newblock Survey and critique of techniques for extracting rules from trained
  artificial neural networks.
\newblock \emph{Knowledge-based systems}, 8\penalty0 (6):\penalty0 373--389,
  1995.

\bibitem[Hinton et~al.(2015)Hinton, Vinyals, and Dean]{hinton2015}
Geoffrey Hinton, Oriol Vinyals, and Jeffrey Dean.
\newblock Distilling the knowledge in a neural network.
\newblock In \emph{NIPS Deep Learning and Representation Learning Workshop},
  2015.
\newblock URL \url{http://arxiv.org/abs/1503.02531}.

\bibitem[Hinton and Frosst(2017)]{frosst2017}
Geoffrey Hinton and Nicholas Frosst.
\newblock Distilling a neural network into a soft decision tree.
\newblock 2017.
\newblock URL \url{https://arxiv.org/pdf/1711.09784.pdf}.

\bibitem[Ribeiro et~al.(2016)Ribeiro, Singh, and Guestrin]{ribeiro2016}
Marco~Tulio Ribeiro, Sameer Singh, and Carlos Guestrin.
\newblock " why should i trust you?" explaining the predictions of any
  classifier.
\newblock In \emph{Proceedings of the 22nd ACM SIGKDD international conference
  on knowledge discovery and data mining}, pages 1135--1144, 2016.

\bibitem[Barocas and Selbst(2016)]{barocas2016}
Solon Barocas and Andrew~D Selbst.
\newblock Big data's disparate impact.
\newblock \emph{California Law Review}, 104:\penalty0 671--732, 2016.

\bibitem[Burrell(2016)]{burrell2016}
Jenna Burrell.
\newblock How the machine ‘thinks’: Understanding opacity in machine
  learning algorithms.
\newblock \emph{Big Data \& Society}, 3\penalty0 (1), 2016.

\bibitem[Stilgoe(2018)]{stilgoe2018}
Jack Stilgoe.
\newblock Machine learning, social learning and the governance of self-driving
  cars.
\newblock \emph{Social studies of science}, 48\penalty0 (1):\penalty0 25--56,
  2018.

\bibitem[Szegedy et~al.(2014)Szegedy, Zaremba, Sutskever, Bruna, Erhan,
  Goodfellow, and Fergus]{szegedy2014}
Christian Szegedy, Wojciech Zaremba, Ilya Sutskever, Joan Bruna, Dumitru Erhan,
  Ian Goodfellow, and Rob Fergus.
\newblock Intriguing properties of neural networks.
\newblock In \emph{International Conference on Learning Representations}, 2014.

\bibitem[Nguyen et~al.(2015)Nguyen, Yosinski, and Clune]{nguyen2015}
Anh Nguyen, Jason Yosinski, and Jeff Clune.
\newblock Deep neural networks are easily fooled: High confidence predictions
  for unrecognizable images.
\newblock In \emph{Proceedings of the IEEE Conference on Computer Vision and
  Pattern Recognition}, pages 427--436, 2015.

\bibitem[Moosavi-Dezfooli et~al.(2016)Moosavi-Dezfooli, Fawzi, and
  Frossard]{moosavi-dezfooli2016}
Seyed-Mohsen Moosavi-Dezfooli, Alhussein Fawzi, and Pascal Frossard.
\newblock Deepfool: a simple and accurate method to fool deep neural networks.
\newblock In \emph{Proceedings of the IEEE Conference on Computer Vision and
  Pattern Recognition}, pages 2574--2582, 2016.

\bibitem[Su et~al.(2019)Su, Vargas, and Sakurai]{su2019}
Jiawei Su, Danilo~Vasconcellos Vargas, and Kouichi Sakurai.
\newblock One pixel attack for fooling deep neural networks.
\newblock \emph{IEEE Transactions on Evolutionary Computation}, 2019.

\bibitem[Alzantot et~al.(2018)Alzantot, Balaji, and Srivastava]{alzantot2018}
Moustafa Alzantot, Bharathan Balaji, and Mani Srivastava.
\newblock Did you hear that? adversarial examples against automatic speech
  recognition.
\newblock \emph{arXiv preprint arXiv:1801.00554}, 2018.

\bibitem[Papernot et~al.(2016)Papernot, McDaniel, Swami, and
  Harang]{papernot2016}
Nicolas Papernot, Patrick McDaniel, Ananthram Swami, and Richard Harang.
\newblock Crafting adversarial input sequences for recurrent neural networks.
\newblock In \emph{MILCOM 2016-2016 IEEE Military Communications Conference},
  pages 49--54. IEEE, 2016.

\bibitem[Kurakin et~al.(2016)Kurakin, Goodfellow, and Bengio]{kurakin2016}
Alexey Kurakin, Ian Goodfellow, and Samy Bengio.
\newblock Adversarial examples in the physical world.
\newblock \emph{arXiv preprint arXiv:1607.02533}, 2016.

\bibitem[Brown et~al.(2017)Brown, Mane, Roy, Abadi, and Gilmer]{brown2017}
Tom Brown, Dandelion Mane, Aurko Roy, Martin Abadi, and Justin Gilmer.
\newblock Adversarial patch.
\newblock 2017.
\newblock URL \url{https://arxiv.org/pdf/1712.09665.pdf}.

\bibitem[Jan et~al.(2019)Jan, Messou, Lin, Huang, and Wang]{jan2019}
Steve~TK Jan, Joseph Messou, Yen-Chen Lin, Jia-Bin Huang, and Gang Wang.
\newblock Connecting the digital and physical world: Improving the robustness
  of adversarial attacks.
\newblock 2019.

\bibitem[Erhan et~al.(2009)Erhan, Bengio, Courville, and Vincent]{erhan2009}
Dumitru Erhan, Yoshua Bengio, Aaron Courville, and Pascal Vincent.
\newblock Visualizing higher-layer features of a deep network.
\newblock \emph{Technical Report 1341}, 1341:\penalty0 1--13, 2009.

\bibitem[Simonyan et~al.(2014)Simonyan, Vedaldi, and Zisserman]{simonyan2014}
Karen Simonyan, Andrea Vedaldi, and Andrew Zisserman.
\newblock Deep inside convolutional networks: Visualising image classification
  models and saliency maps.
\newblock In \emph{ICLR}, 2014.

\bibitem[Fong and Vedaldi(2017)]{fong2017}
Ruth~C Fong and Andrea Vedaldi.
\newblock Interpretable explanations of black boxes by meaningful perturbation.
\newblock In \emph{Proceedings of the IEEE International Conference on Computer
  Vision}, pages 3429--3437, 2017.

\bibitem[Le(2013)]{le2013}
Quoc~V Le.
\newblock Building high-level features using large scale unsupervised learning.
\newblock In \emph{2013 IEEE International Conference on Acoustics, Speech and
  Signal Processing (ICASSP)}, pages 8595--8598. IEEE, 2013.

\bibitem[Karpathy et~al.(2016)Karpathy, Johnson, and Fei-Fei]{karpathy2016}
Andrej Karpathy, Justin Johnson, and Li~Fei-Fei.
\newblock Visualizing and understanding recurrent networks.
\newblock \emph{ICLR Workshop}, 2016.

\bibitem[Olah et~al.(2018)Olah, Satyanarayan, Johnson, Carter, Schubert, Ye,
  and Mordvintsev]{olah2018}
Chris Olah, Arvind Satyanarayan, Ian Johnson, Shan Carter, Ludwig Schubert,
  Katherine Ye, and Alexander Mordvintsev.
\newblock The building blocks of interpretability.
\newblock \emph{Distill}, 2018.
\newblock \doi{10.23915/distill.00010}.
\newblock https://distill.pub/2018/building-blocks.

\end{thebibliography}

\end{document}